\documentclass[journal]{IEEEtran}
\usepackage{graphicx}
\DeclareGraphicsExtensions{.pdf,.jpg,.png,.eps}
\usepackage{amsmath,amssymb}
\usepackage{color}
\newcommand{\etal}{\textit{et al}. }
\newcommand{\ie}{\textit{i}.\textit{e}. }
\newcommand{\eg}{\textit{e}.\textit{g}. }

\usepackage{xcolor}
\hypersetup{
    colorlinks,
    linkcolor={red},
    citecolor={blue},
    urlcolor={blue}
}
\usepackage{color, colortbl}
\definecolor{Gray}{gray}{0.9}
\usepackage{authblk}

\begin{document}

\title{TrackNet: Simultaneous Object Detection and Tracking and Its Application in Traffic Video Analysis}

\author{\IEEEauthorblockN{Chenge Li\IEEEauthorrefmark{1}, Gregory Dobler\IEEEauthorrefmark{1}, Xin Feng\IEEEauthorrefmark{2}, Yao Wang\IEEEauthorrefmark{1}}\\
\IEEEauthorrefmark{1}\IEEEauthorblockA{Department of Electrical and Computer Engineering, Tandon School of Engineering\\
New York University, Brooklyn, NY 11201, USA\\
\{cl2840, greg.dobler, yw523\}@nyu.edu}\\
\IEEEauthorrefmark{2}\IEEEauthorblockA{
Chongqing University of Technology, Chongqing, China\\
\{fx\_328\}@hotmail.com}
}

\maketitle

\begin{abstract} Object detection and object tracking are usually treated as two separate processes. Significant progress has been made for object detection in 2D images using deep learning networks. The usual ``tracking-by-detection" pipeline for object tracking requires that the object is successfully detected in the first frame and all subsequent frames, and tracking is done by ``associating'' detection results. Performing object detection and object tracking through a single network remains a challenging open question. We propose a novel network structure named \textbf{trackNet} that can directly detect a 3D tube enclosing a moving object in a video segment by extending the faster R-CNN framework. A \textbf{Tube Proposal Network (TPN)} inside the trackNet is proposed to predict the objectness of each candidate tube and location parameters specifying the bounding tube. The proposed framework is applicable for detecting and tracking any object and in this paper, we focus on its application for traffic video analysis. The proposed model is trained and tested on UA-DETRAC, a large traffic video dataset available for multi-vehicle detection and tracking, and obtained very promising results.
\end{abstract}


\section{Introduction}

\IEEEPARstart{O}{bject} 
 detection and object tracking have been two longstanding challenges for the computer vision community, and much progress has been made on both fronts. For object detection, complex hand-crafted features plus shallow classifiers such as HOG+SVM \cite{hog} and multiple resolution image pyramids plus multiple filters such as DPM \cite{dpm} were both popular detection pipelines. In recent years, Convolutional Neural Networks(CNN) have enjoyed great development and CNN based methods such as \cite{he2017mask,rcnn,faster_rcnn,ssd,yolo9000,yolo} have been setting up new records in object detections in still images. Object tracking, especially Multi-Object Tracking (MOT), has many real-world applications including intelligent transportation\cite{luo2018fast, shao2018seaships}, virtual/augmented reality, robot navigation, etc. Most existing MOT systems can be classified into two groups \cite{mot_review}: Detection Based Tracking (DBT) and Detection Free Tracking (DFT). DBT requires object being detected for every frame followed by a tracker that links the detection regions based either on object features or probabilistic movement characteristics. The performance of DBT highly depends on the performance of the employed object detector. DFT on the other hand, requires manual initialization of objects in the initial frame and couldn't handle uninitialized objects.

\begin{figure*}[t]
\begin{center}
  \includegraphics[width=1\textwidth]{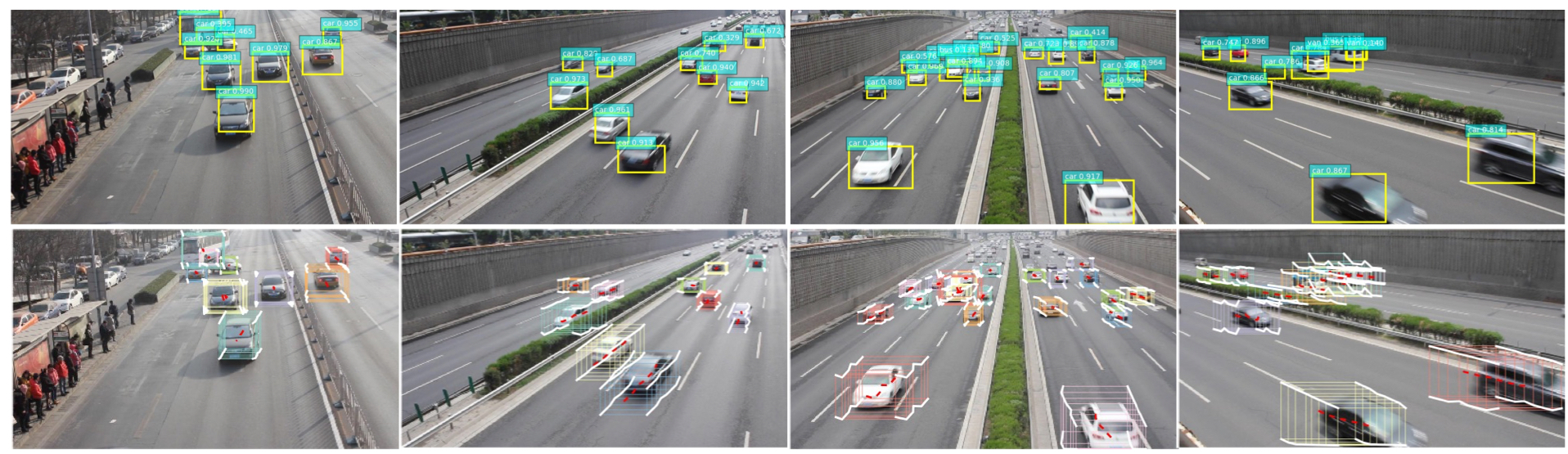}
   \caption{Upper row(the current 2D object detection results): bounding boxes in a certain frame. Lower row(our proposal results): bounding tubes for consecutive $T$ frames. Moving vehicles in short time periods possess smooth motion. Generating spatial-temporal bounding tubes is a more efficient way to model objects' motion than associating per-frame bounding boxes.}
\label{teaser}
\end{center}
\end{figure*}

We argue that object detection and tracking should not be treated as two independent tasks, but rather, effective object detection in video should employ both spatial appearance and temporal motion features.  In this work, we propose a unified network structure for simultaneous object detection and tracking. Our network treats a group of consecutive pictures (GoP) as a 3D volume, and detects moving objects in the GoP as tubes within it.

\subsection{Motivations for Tube Proposals} 
Bounding box proposal generation is an essential step for most state-of-the-art object detection systems \cite{he2017mask,rcnn,faster_rcnn} (excluding proposal-free detectors\cite{ssd,yolo}). What's more, \cite{tang2017object} proposed to combine multiple object proposals from different proposal generation methods such as Selective Search\cite{selective_search}, Edge Boxes\cite{zitnick2014edge}, and RPN\cite{faster_rcnn} to boost the recall performance. \cite{li2018multistage} proposed to recursively refine object proposals with multiple iterations. In this work, we propose to use tube proposals instead of box proposals. One obvious advantage of tube proposals over box proposals is the convenience of getting all objects' spatial-temporal locations in one shot. A bounding tube is readily available after one forward pass instead of $N$ times of forward passes plus post-processing association. Secondly, instead of extracting spatial contextual information using separate subnetworks\cite{li2017attentive}, tube proposals on multiple consecutive frames naturally provides global and local context in the spatial-temporal domain. The other advantage of tube proposals is its role to provide a much stronger regularization during training. Suppose that there are $B$ moving objects, which appeared in each of the $N$ frames. If we first detect $B$ box proposals in each frame, we would have to examine $B^N$ possible tubes, while there are only $B$ tubes that are correct. Ground truth bounding tubes occur very rarely in this high-dimensional space, and each one of them carries highly structured information implicitly. This sparsity and implicit structure information will serve as very strong regularization: only tubes with certain spatial and motion features over the entire GoP are good candidates. This motion pattern regularization is even more obvious when dealing with traffic videos, as shown in figure \ref{teaser}. Therefore, the step from proposing boxes to proposing tubes is a very natural extension.

\section{Related Work}
\textbf{Object Detection in Images}. Object detection in images has progressed rapidly in recent years \cite{rcnn,faster_rcnn,ssd,yolo9000,yolo}. Girshick \etal first introduced R-CNN \cite{rcnn} to identify and label objects in 2D images. From object region proposals that are generated by an independent algorithm (e.g., selective search \cite{selective_search}, Edge Boxes \cite{zitnick2014edge}), R-CNN runs a forward pass once for each proposed region to determine whether this region contains an object. The same authors further improved R-CNN to fast R-CNN\cite{fast_rcnn} by sharing convolution feature maps among all object region proposals, and hence only one forward pass for all region proposals is needed. Faster R-CNN \cite{faster_rcnn} further improved upon fast R-CNN by introducing a region proposal network (RPN) that directly regresses fixed ``anchors'' to object region proposals from feature maps extracted using a 2D convolutional network. 

\textbf{Object Detection in Videos}. Most systems proposed so far for identifying (and tracking) moving objects in videos rely on 2D object detection in each frame, which is computationally expensive and does not jointly consider object(s') motion information. Some systems (e.g., \cite{t-CNN,tubelet}) have used explicit motion information (\eg optical flow) as a linking feature to associate detection regions or to smooth detection scores as a post-processing step. Those motion information is derived separately outside of the network and is not integrated organically with the network training. Inspired by the correlation and regression based trackers such as \cite{bertinetto2016fully,held2016learning,ma2015hierarchical,chen2018learning}, \cite{feichtenhofer2017detect} proposed a D\&T framework which relies on a resNet-101 as frame-level feature extractor and two parallel region proposal networks (RPN) to generate 2D box proposals. The detected boxes are associated using a proposed ROI tracking module by computing the correlation map between the feature maps. Another work in \cite{lu2017online} first uses SSD\cite{ssd} to detect objects and extracts corresponding spatial features through ROI pooling to create per-frame feature. Then an association LSTM is proposed to regress and associate object locations given the frame-level feature tensor for past $\tau$ consecutive frames as input.

\begin{figure*}[t]
\begin{center}
  \includegraphics[width=1.0\textwidth]{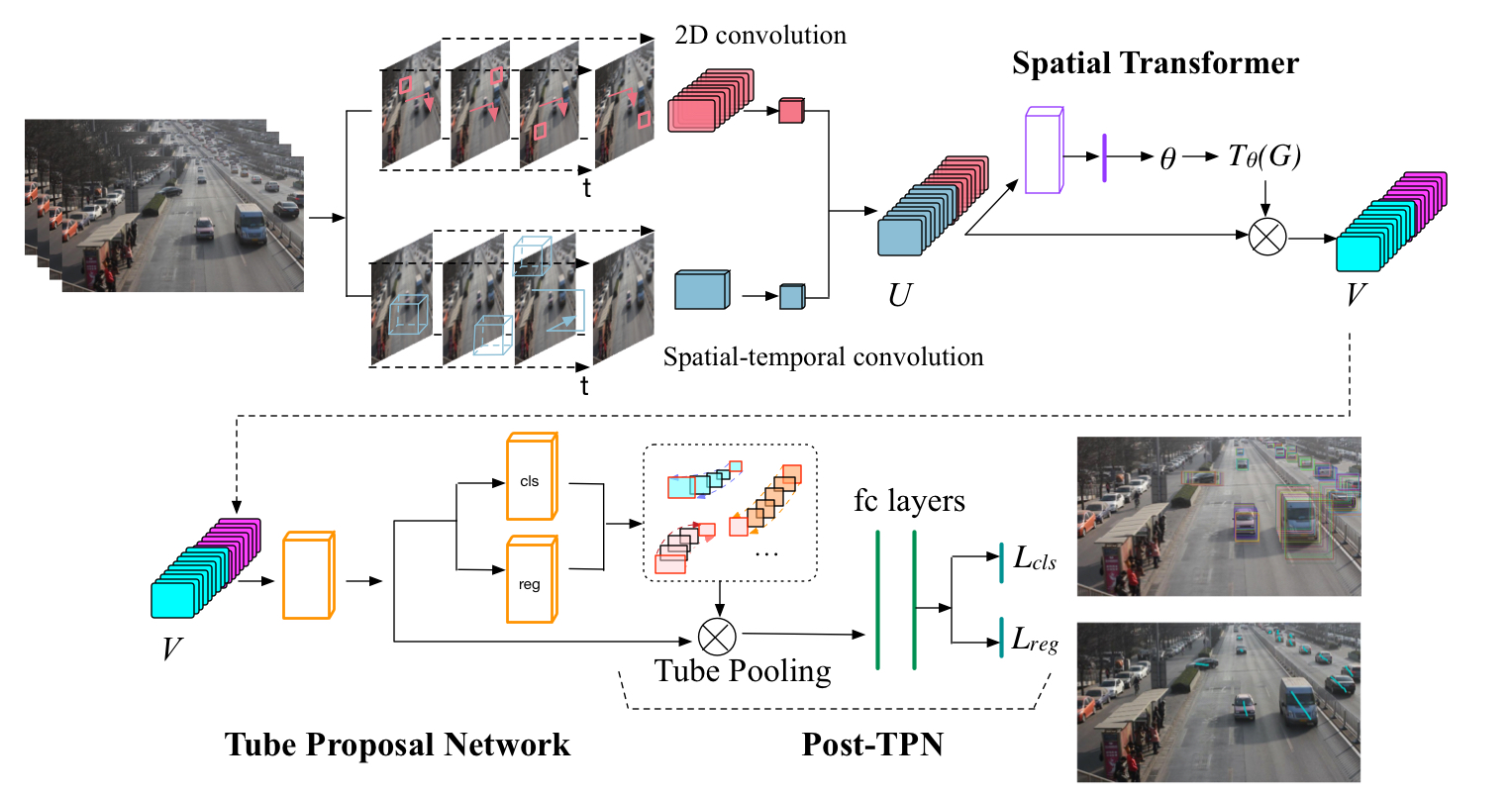}
   \caption{The trackNet structure. A two branched backbone produces both appearance-only and spatial-temporal features from a video GoP. A spatial transformer\cite{transformer} is inserted to introduce global feature warping to a consistent viewing angle. A tube proposal network is introduced to generate flexible bounding tube proposals. Proposal tubes are used to guide more specific pooling, and the tube-pooled features are used to further predict more specific class labels and refine tube locations. The final output will be bounding tubes for moving objects inside this video GoP. A short tracklet is also generated by connecting the centroids of bounding boxes at each frame.}
\label{flowchart}
\end{center}
\end{figure*}

\textbf{Tube Proposal Based Works}. A more related branch of works with this paper is the \textit{tube proposal} based methods such as \cite{tpn,saha2017amtnet,hou2017tube,kalogeiton2017action}. In all these works, the initial anchor tubes are generated from duplicating the same bounding box (anchor) across multiple frames. Such a tube describes a non-moving object and will be called ``stationary tubes". Spatial per-frame features are then pooled from the same box location across multiple frames. The 3D region proposals in \cite{saha2017amtnet} are two-frame micro-tubes, which are a pair of bounding boxes spanning 2 video frames separated by a predefined temporal interval $\Delta$ ($\Delta$=1 or 2), which are almost consecutive frames. During training, the intersection-over-union (IoU) and regression loss are computed between the ground truth pair and the proposal pair. Whereas in our implementation, GoP length $T$ can be varying and much longer than 2 frames, \eg 8 or 16. What's more, our network uses a loss function that considers the difference between box positions and the ground truth locations in every intermediate frame. In terms of features, \cite{saha2017amtnet} uses spatial feature fusion from VGG feature maps only for the front and ending frames. \cite{kalogeiton2017action} fuses spatial and motion information from two streams: a SSD\cite{ssd} as the appearance detector and a motion detector taking the optical flow images as input.

The major limitations of the above prior work are two-fold. Firstly, stationary anchor tubes correspond to objects with no motion in $N$ frames, which tend to have very low overlap with ground truth tubes covering moving objects. This has serious negative impact on the network training: Only those anchor tubes with sufficient overlap with ground truth tubes can be used for tuning the tube offset regression module, and using stationary anchors significantly reduces such positive anchor tubes. Second, models which pool only spatial features do not explicitly exploit motion information, which is an important cue for moving object detection and tracking.  Our proposed network differs from these prior work in the following aspects: 1) Our model initializes anchor tubes based on the motion vectors near the anchor location to allow non-stationary tubes as starters. 2) Our end-to-end network captures the spatial and temporal information simultaneously over all frames using both 2D and 3D convolutional neural networks for feature extraction, which does not require a separate input such as the flow images as in \cite{kalogeiton2017action}. It is faster than fusing features from CNN and features from other techniques, and simpler than using LSTM to learn the feature associations/correlations. 3) In the tube refinement stage, 3D (as opposed to 2D) tube-ROI pooling enables features from different box locations to be pooled. 4) The loss function for the tube position regression considers the ground truth positions of the object in all frames, even when we parameterize the tube position using linear interpolation from two bounding boxes in the beginning and ending frames. Thus the detected final tube maximizes the overlap with the ground-truth object over all frames. This is particularly important when the underlying object does not follow a linear trajectory.


\section{TrackNet Model Architecture}

\subsection{Two Stream Feature Extraction and Transformation} 
In order to utilize both spatial and temporal features, our network is based on VGG net trained from ImageNet and C3D trained from UCF101 for video classifications. Figure \ref{flowchart} provides an overview of the trackNet structure. We divide a video into group of pictures (GoP) of fixed length $T$ (8 frames in our implementation) and feed the raw video frames in each GoP into a two-stream backbone structure, where the first branch is a VGG16 subnetwork with convolutions and spatial max-poolings and the second branch is a C3D-like subnetwork with 3D convolutions and spatial-temporal max-poolings. These two kinds of features compliment each other in that one focuses more on appearance whereas the other focuses more on motions. The resulting 2D feature maps ($T\times h\times w \times 512 $, reshape to $1\times h\times w \times 512T $) and the spatial-temporal feature maps ($1\times \frac{T}{8}\times h\times w \times 512 $) then separately go through a ``squashing" convolutional layer (bottleneck layer) with $1\times1$ kernel size, which reduces the number of feature maps to 128 for each stream. Squashed feature maps are concatenated afterwards. 

\textbf{Spatial Transformer}. 

\begin{figure}[t]
\begin{center}
 \includegraphics[width=1.0\linewidth]{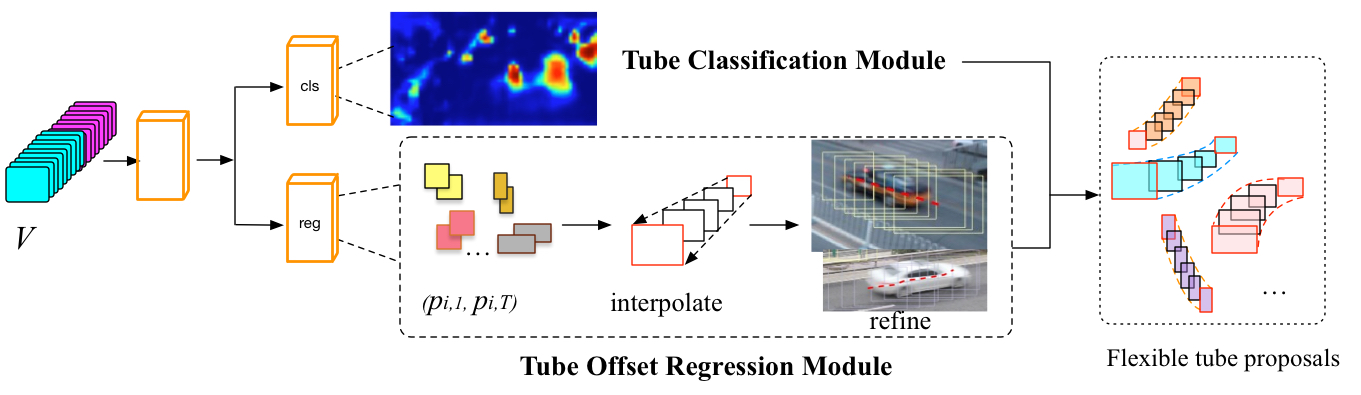}
   \caption{Tube Proposal Network (TPN) in detail: based on the shared convolution feature maps, a TPN consists of two parts: the classification module and the tube offset regression module. We show a heat map of the predicted objectness scores from the classification module: regions where objects are moving have higher (warmer) values.}
\label{TPN}
\end{center}
\end{figure}

When dealing with real-world videos, sometimes the network will observe objects' frontal appearance, whereas at other times, the side appearances of objects will be observed. Inspired by \cite{transformer}, we utilized a learnable module, the spatial transformer, to map the concatenated features from different viewing angles into a ``unified" manifold. We used affine transformation in our case, however, one can use more complicated transformations to suit their cases as indicated in \cite{transformer}. Our transformer has a very simple structure with only one convolutional layer and one fully connected layer as shown in figure \ref{flowchart}. Six affine transformation parameters $\theta$ will be output from the fully connected layer and then used to transform the original concatenated feature maps $U$. Instead of sampling from the original feature maps using a regular mesh grid $G$, the sampling grid will be transformed using $\theta$ to $T_\theta(G)$, which is applied to original feature maps $U$ to produce the warped output feature maps $V$. All feature maps are transformed channel-wise in the same way. The transformed features $V$ are then fed into the tube proposal network (TPN) shown in figure \ref{TPN} to generate tube proposals and a post-TPN stage to further refine those proposals. We will describe the details of different subcomponents in the following sections.

\subsection{Tube Proposal Network (TPN)} 
TPN will produce many initial tube proposals. Similar to Faster R-CNN's region proposal network (RPN), the TPN generates multiple candidate anchor tubes at each pixel location. For each anchor tube, the classification module predicts the objectness score of the tube (\ie probability of having objects inside the tube) and the offset regression module generates the position offsets from the anchor tubes to the ground truth bounding tubes. Both the classification module and offset regression module inside TPN are sharing the same transformed feature maps $V$ of size $H\times W \times 256$. 

\textbf{Stationary and Tilted Anchor Tubes.} 
Faster R-CNN's RPN uses $M$ predefined anchors with different base sizes and aspect ratios. Analogously, we start from a fixed set of anchor tubes.  As discussed in the introduction section, the tube space is a very high-dimensional space. It is not possible and not necessary to consider every possible tube. Consider a short video segment with $T$ frames (\eg 8 frames in an $FPS=30$ video), an object's trajectory is usually very smooth and nearly linear in such a short time period. These quasi-linear tubelets may differ in size, direction or speed, however they all live on a low dimensional manifold with limited degrees of freedom. Because of the quasi-linearity of objects' short trajectories, we use straight anchor tubes as the initial candidates. A naive way to construct an anchor tube is to have the same bounding box positions in all frames\cite{tpn,saha2017amtnet,hou2017tube,kalogeiton2017action}, which we call ``stationary tubes" $T_s$, as they correspond to non-moving objects. 

These simple tubes, however, are not always desirable, especially when dealing with videos with varying viewing points and camera positions. Consider the typical scenes from the UA-DETRAC\cite{uadetrac} dataset (see sample pictures in figure \ref{tubes}), traffic can flow towards an arbitrary direction, and a stationary tube would have very low overlapping with the true tube, resulting in a very big offset, which is hard to regress. In order to have better initial tube candidates, we utilize motion vectors (MV) derived from optical flow fields and construct tilted tubes by modifying the initial stationary tubes using the average motion directions. Firstly the dominant motion direction (positive or negative) will be decided based on the votes of all the motion vectors at this pixel location in this video GoP. If the dominant direction is positive, then the first half of the biggest motion vectors will be averaged as the mean MV, while if the dominant direction is negative, then the first half of the smallest motion vectors will be averaged as the result. For frame index $t=[1,N]$, box positions of the tilted tube $T_t$ can be derived from the straight tubes $T_s$ as: $T_t[t, :,:,k] = T_s[t, :,:,k]+mv\times (t-1)$, where $mv$ is either $mv_x$ or $mv_y$, the mean motion vectors in $x$ and $y$ directions at each pixel location. $(T[t, w,h,0],T[t, w,h,1])$ is the upper left corner position whereas $(T[t, w,h,2],T[t, w,h,3])$ is the lower right corner position of the bounding box at $(w,h)$ in frame $t$.

Unlike faster R-CNN, we do not pre-define the size and aspect ratio for the anchors. Rather we identify the typical sizes and aspect ratios of the ground truth bounding boxes in the training data using the K-means clustering (inspired by YOLO \cite{yolo}), and use the centroids for all clusters as the sizes and aspect ratios for the base stationary anchors. We find that the network has difficulty regressing to large offset values if starting from poorly chosen initial anchors and therefore an appropriate initialization for anchor sizes, aspect ratios and moving directions is rather important.

\begin{figure}
\begin{center}
\includegraphics[width=0.9\linewidth]{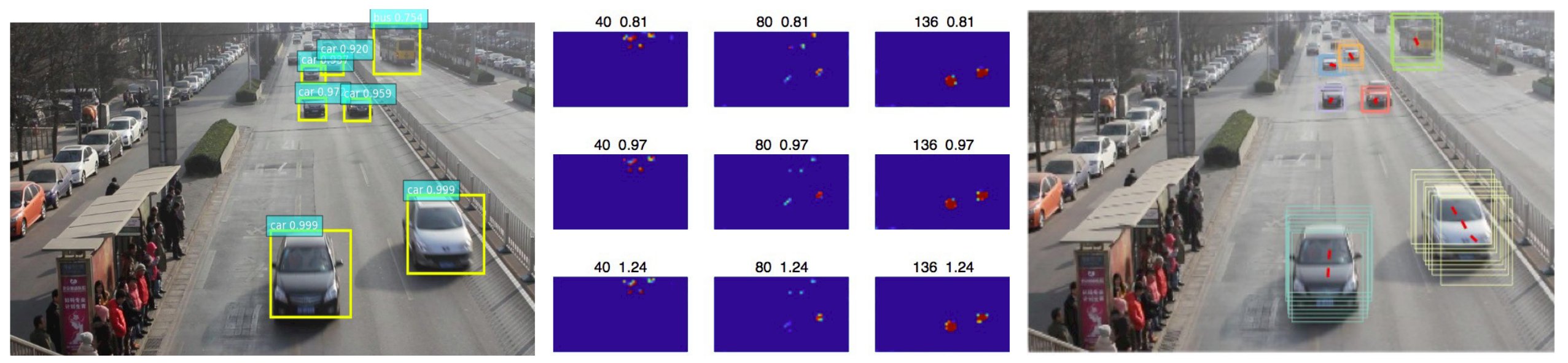}
\caption{Example of the $M=9$ objectness scores shown in the middle as the heat maps. Each score heat map corresponds to a specific set of anchor size and aspect ratio. $\eg$ the top left one corresponds to size=40, aspect ratio=0.81. It can be seen that score maps of smaller-sized anchor tubes are more sensitive to smaller objects whereas score maps of larger-sized anchor tubes are more sensitive to larger objects.}
\label{withheatmap}
\end{center}
\end{figure}

\textbf{Tube Classification Module.}
Based on the shared feature maps $V$, $M$ objectness scores are predicted in the classification module for $M$ anchor tubes at each feature map location by a convolutional layer with kernel size $3\times 3$ (acting on $K=256$ feature maps). Essentially the objectness score for each anchor is determined from a $3\times 3\times K$ feature tensor. This module can be viewed as a fully convolutional network. During training, tube overlapping, \ie 3D intersection over union (3D-IoU) between the anchor tubes and ground truth tubes are computed. Anchor tubes with high 3D-IoU will be selected as positive proposals and assigned label $+1$, whereas anchor tubes with low 3D-IoU scores (partially overlapped) will be assigned label $-1$ and the remaining anchor tubes (including pure background) will be ignored. The classification module is trained with the cross-entropy classification loss $L_{cls_{TPN}}$ with respect to their ground truth label. In figure \ref{TPN}, the tube classification module is shown. The heat map is the objectness score output, where bigger (warmer) values indicate higher probabilities of containing objects in that location. Here we only show one heat map corresponding to one set of anchor tube size. In our implementation, we have $M=9$ set of base tubes with different sizes or aspect ratios as in figure \ref{withheatmap}.

\textbf{Tube Offset Regression Module.}
\label{tpn_reg}
Anchor tubes will be ranked based on their objectness scores. For tubes with higher scores, the offsets between the corner positions of the tubes and ground truth positions are computed as the regression targets. The regression module will be trained to generate these regression targets from the input $3\times 3\times K$ features. 
Given $M$ candidate anchor tubes at each feature map location, the offsets will be predicted so as to ``bend'' the straight anchor tubes into a shape closer to the ground truth tubes. Following R-CNN,  we use the center position and width and height to parameterize the position of a rectangular bounding box in each frame. The offsets of these parameters between the bounding boxes of all frames in an anchor tube (ST) and those in the ground truth tube (GT) are our \textbf{3D tube regression target} for this anchor tube. We adopt the parameterization of the 4 coordinates in \cite{rcnn}, but similar as \cite{yolo}, we normalize the spatial coordinate by the actual width and height of the video frame, so that the normalized coordinate and hence the 4 parameters are all in the range of $[0,1]$, which helps the convergence speed. The 3D Tube regression targets for positive anchor tube $i$ at frame $t$ is defined as:

$tar_{i,t}=$
\begin{equation*}
\fontsize{8}{10}
\begin{bmatrix}
\Delta X_{t}^{gt} \\
\Delta Y_{t}^{gt} \\
\Delta W_{t}^{gt} \\
\Delta H_{t}^{gt}
\end{bmatrix}
=
\begin{bmatrix}
\frac { (GT_{center \ x})_t -{(ST_{center \ x})}_t }{(ST_w)_t} \\
\frac { {(GT_{center \ y})}_t -{(ST_{center \ y})}_t}{(ST_h)_t} \\
log \frac { (GT_w)_t}{ (ST_w)_t}\\
log \frac { (GT_h)_t}{(ST_h)_t}
\end{bmatrix}
\end{equation*}

By learning to regress to these targets, the system can derive the refined locations for all anchor tubes that have high overlap with ground truth bounding tubes. We have explored two ways to wire the tube offset regression module: $(1)$ directly predicting offsets of all frames and $(2)$ utilizing linear interpolation. 

\textbf{Option 1: Directly predict tube parameters}. In this structure, we directly estimate the offsets of every frame. Given a video GoP of length $T$, the regression network directly predicts $4\times T$ parameters for every tube. As our straight tube candidates are spreading over all pixel locations, the regression network is implemented using a convolution layer with $4\times T\times M$ output maps.

\textbf{Option 2: Linear interpolation of bounding box offsets from offsets at two frames}. Despite the fact that an object inside a video can have arbitrarily complex motions, most objects' motions are very smooth in real-world videos. Given a short enough time period, we can approximate the trajectory of each corner of the bounding tube with a straight line. This is particularly true for traffic videos containing moving vehicles. Motivated by this observation, instead of determining the offsets of the corner positions in all frames, the regression network only estimates the offsets in the beginning and ending frames, and linearly interpolate the offsets in other frames. During training, the regression loss considers the difference between the true offsets (targets) and the estimated offsets for all frames, which are interpolated from the offsets in the beginning and ending frames. The advantage of this approach is that only $8$ parameters are estimated for a given tube, as opposed to $4\times T$ parameters. Compared to directly estimating the offset at every frame, this approach also implicitly applies a smoothness constraint along the corner trajectories and prevents the network to generate erratic trajectories. 

We implement the interpolation using a convolution layer with spatial $1\times 1$ kernel. For example, if we have a video segment with length $T=8$ frames, $\Delta X_1$, $\Delta Y_1$, $\Delta W_1$, $\Delta H_1$, $\Delta X_{T}$, $\Delta Y_{T}$, $\Delta W_{T}$, $\Delta H_{T}$ are the predicted center offsets and width and height offsets at the first frame $(t=1)$ and the last frame $(t=8)$. The offset at time frame $t$ can be easily implemented using a convolutional layer with $1\times1 \times 2 \times 8$ kernel matrix: \\
\begin{equation}
\fontsize{8}{10}
K=
  \begin{bmatrix}
1,6/7,  5/7, 4/7, 3/7, 2/7, 1/7, 0\\
0, 1/7, 2/7, 3/7, 4/7, 5/7, 6/7, 1\\ 
\end{bmatrix}
\end{equation}

If we view the first frame prediction result (with 4 channels for 4 parameters) and the last frame prediction result as $2$ separate input feature maps, then these $2$ feature maps convolving with this $1\times 1 \times 2 \times 8$ kernel will produce $8$ feature maps, corresponding to predicted offsets for all $8$ frames. Note that we could implement higher order interpolation by using more than 2 input feature maps and setting the kernel matrix accordingly.  We could also train the kernel matrix as part of the regression network to learn the appropriate interpolation kernel.

We adopted option 2 (linear interpolation) for tube proposals during TPN stage to save parameters and constrain smooth motions, and relaxed to option 1 (predict all) in the post-TPN stage to further refine locations.

\subsection{post-TPN: Classification and Refinement}
As shown in Figure \ref{flowchart}, the tube proposal network generates many tube proposals, whose positions are determined by the original candidate tubes and the predicted offsets. Proposal tubes with high objectness scores will go through a second stage of classification and regression. In this stage, tube proposals will be further classified into different classes (such as car, bus, van etc. for UA-DETRAC dataset). The position offsets for the tube will also be refined. Instead of using the features pooled from the $3\times3$ neighborhood on the feature map as in TPN, features specific for the proposal tube regions are pooled using the tube pooling. 

\textbf{Tube Pooling}. ROI pooling was introduced in \cite{spp}, which enables different proposal regions to be described by the same dimensional feature vectors. In our case, a proposal tube consists of bounding boxes in different frames that are different in sizes and locations. Pooling based on one particular bounding box inside the tube would be deficient. Instead of pooling from the same ROI location multiple times as in \cite{tpn}, the union of all bounding boxes in a proposal tube is found and features covering the union region are extracted from the transformed feature maps $V$. After the tube pooling, this feature vector is then fed into a post-TPN subnetwork, which further assesses its class and refines the tube position information.

 There are two fully connected (fc) layers and another two fc layers for predicting classification scores and offsets separately. In our implementation, $256$ proposal tubes (half positive, half negative) are considered and $7\times7\times 256$ features are pooled from feature map $V$ using the tube union ROI, leading to a total of $256 \times 7\times 7 \times 256$ features. For the offset regression, similar to the TPN regression module, either linear interpolation or directly predicting offsets at all frames can be chosen.

\subsection{Multi-task Loss to Train the TrackNet}
Both classification loss and regression loss are used to penalize the proposed tubes. For the TPN, the predicted objectness score for each anchor tube will have the cross-entropy classification loss $L_{cls_{TPN}}$ with respect to their ground truth label. Positive anchor tubes will have the regression loss $L_{reg_{TPN}}$ with respect to the offset targets. During the post-TPN stage, true class labels (\eg background, car, bus, van) are used for the cross-entropy loss $L_{cls}$. Both regression losses $L_{reg_{TPN}}$ and $L_{reg}$ use the smooth  $l_1$ loss defined in \cite{fast_rcnn}.

The above losses are combined to form the total loss for a proposal tube:
$L(s_i,p_{i}) = $
\begin{equation}
\begin{aligned}
\lambda_1 \times  L_{cls}(l_i, s_i)  
+\lambda_2 \times  \sum\limits_{t=1}^{T} L_{reg}(tar_{i,t}, p_{i,t}) \\
 +\lambda_3 \times  L_{cls_{TPN}}(l_i, s_i) 
 + \lambda_4 \times  \sum\limits_{t=1}^{T} L_{reg_{TPN}}(tar_{i,t}, p_{i,t})\\
  +\lambda_5 \times L_{smooth} \\
\end{aligned}
\end{equation}

where $l_i$ is the ground truth label for anchor tube $i$, $s_i$ is the predicted objectness score or the specific class score for anchor tube $i$. $tar_{i,t}$ is the ground truth target, a four-parameter vector representing the offsets between the ground truth location and the location of positive anchor tube $i$ at time $t$, \ie $tar_{i,t}= [\Delta X_{t}^{gt}, \Delta Y_{t}^{gt},\Delta W_{t}^{gt}, \Delta H_{t}^{gt}]_i$. And $p_{i,t}$ is the predicted offset vector, \ie  $p_{i,t}= [\Delta X_{t},\Delta Y_{t},\Delta W_{t},\Delta H_{t}]_i$. When we use the option of directly regressing box locations in each frame, we add a smoothness loss term $L_{smooth}$ to further enhance the smoothness (quasi-linearity) of the tube, which can be derived from the total variation of the tube positions or the average position change between two frames. The $\lambda$s control the weights for different losses. From experiments, we found these hyper parameters are not very sensitive. We set $\lambda_{1,2,3,4}=1$ and $\lambda_5 = 0.001$ in all of the following experiments. The multi-task loss for training is defined as:

\begin{equation}
L_{\mbox{multi-task}}= \sum\limits_{i=1}^{N\_tubes} L(s_i,p_{i}) 
\end{equation}

\section{Experiments}
\textbf{Dataset.} Most of the object detection dataset are 2D images, such as ImageNet \cite{imagenet}, PASCAL VOC  \cite{pascal}, Microsoft COCO \cite{coco}, etc. In the ILSVRC2015 challenge, ImageNet\cite{imagenet} introduced the VID task with 30 categories to attract attention in the object detection in videos. However, most of the videos only contain very few dominant objects, whereas in real world, multiple object detection (MOD) and multiple object tracking (MOT) need to be addressed simultaneously. For example, in traffic analysis and autonomous driving, accurate vehicle detection and tracking, especially in the busy urban area, remains a big challenge. Existing works on vehicle tracking in urban areas such as \cite{li2016robust,jodoin2014urban} utilized traditional methods such as background modeling or feature points tracklet clustering, have limited performances. To evaluate our model regarding both MOD and MOT, we use the UA-DETRAC\cite{uadetrac} dataset, which consists of challenging video sequences captured from real-world traffic scenes with different viewing angles. 

\begin{figure*}[t]
\begin{center}
  \includegraphics[width=1.0\textwidth]{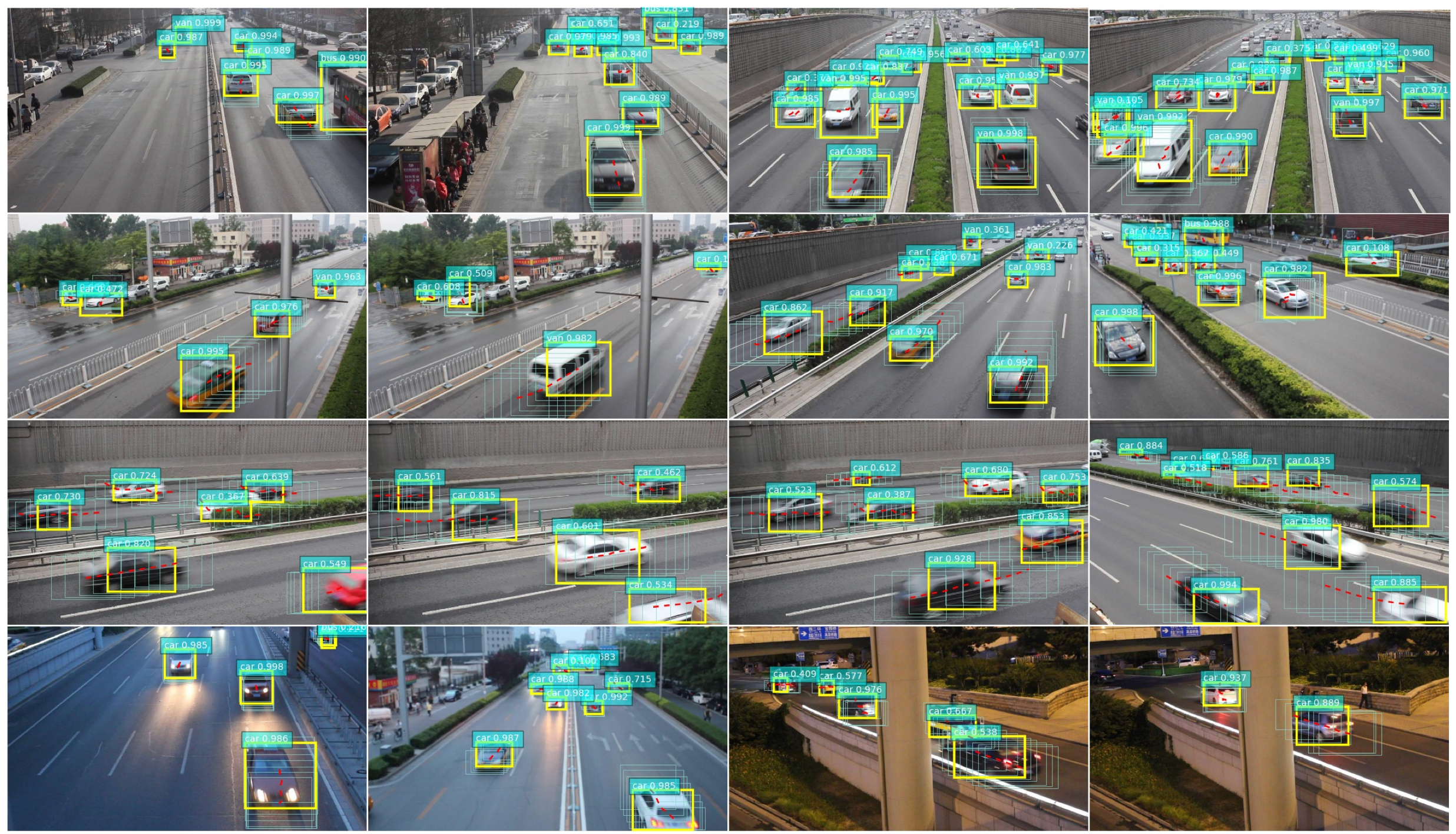}
   \caption{Examples of the predicted bounding tubes. We draw the bounding tube for the entire 8-frame long GoP on the middle frame with the bounding boxes for current frame highlighted in yellow. The centroids of tubes are connected as the tracklets. TrackNet is robust to different lighting conditions and able to cover both small and large vehicle sizes with different aspect ratios. It also generates more ``sparse'' bounding tubes for fast-moving vehicles and ``denser" bounding tubes for slower vehicles or vehicles that are are further away.}
    \label{tubes}
\end{center}
\end{figure*}

We split the dataset into $45$ training and $15$ testing videos and made sure that both training and testing covers all different camera views. The video lengths range from around 700 frames to around 2500 frames. This dataset spans a variety of different weathers such as \textit{sunny}, \textit{cloudy}, \textit{rainy} and \textit{night}. We did not split the dataset to ensure that the training and testing set each includes samples taken under different weather conditions. However, the trained model turns out to be pretty robust to different weather conditions, see figure \ref{tubes}.

\begin{table*}[t]

\begin{center}
\centering
\begin{tabular}{|c| c|c| c| c|c| c| c|c| c| c|}
\hline

&&&& &\multicolumn{3}{ c| }{AP@IoU:} & \multicolumn{3}{ c| }{ AP area:} \\
&$T_\theta(G)$ & VGG &C3D & LP &0.10:1.0 & 0.10 & 0.50 &
small & medium & large \\
\hline

 C3D head only  &&&&&&&&& & \\
 (no VGG, no transformer, predict all) &&&\checkmark& &7.35 & 28.23 & 3.65 & 2.72 & 4.88 & 14.95\\
\hline
  C3D head only w/ LP &&&&&&&&&&\\
 (no VGG, no transformer, interpolate)& &&\checkmark&\checkmark &18.67& 42.50 & 21.55 & 10.55 & 16.45 & 26.33\\
\hline
  TrackNet (no transformer)&&\checkmark&\checkmark&\checkmark  &30.90 & 64.60 & 39.63& 8.99 & 27.71 & 40.01\\
\hline
 TrackNet$^*$ Left view only &  \checkmark&\checkmark&\checkmark&\checkmark 
  &29.28&    62.25&    36.38&   13.97&    24.40&   37.20\\
TrackNet$^*$ Right view only &\checkmark&\checkmark&\checkmark&\checkmark  &26.74 & 57.52 & 33.30 & 5.0 & 21.96 & 33.08\\ 
TrackNet$^*$ Frontal view only &\checkmark&\checkmark&\checkmark&\checkmark  &34.40 & 66.15 & 49.40 & 8.83 & 34.70 & 45.27 \\
\hline
 \rowcolor{Gray}
TrackNet$^*$&\checkmark&\checkmark&\checkmark&\checkmark &31.53 &64.96 & 41.38 & 11.05 & 28.39 &41.29\\
 \rowcolor{Gray}
 TrackNet$^*$(2000 ROI during test) &\checkmark&\checkmark&\checkmark&\checkmark&32.55 & 70.11 & 40.57 & \bf{12.73} & 29.47 & 40.32\\
\rowcolor{Gray}
TrackNet$^*$ (train w/ flipped) &\checkmark &\checkmark&\checkmark&\checkmark & 37.04 & 73.48 & 49.18 & 11.55 & 33.96 & 44.09\\

\rowcolor{Gray}
TrackNet$^*$ (train w/ skip)  & \checkmark&\checkmark&\checkmark&\checkmark &  \bf{37.47}& \bf{73.85} & \bf{50.73} & 11.31 & \bf{34.90} & \bf{44.29}\\
\hline
\textit{\bf{Below: Increase squash dimension from 128 to 512}}:\\
\hline
squashVGG512noC3D & &\checkmark& &\checkmark &35.27 & 71.01 & 46.71 & 11.97 & 31.74 & 42.54 \\
squashVGG512noC3D (train w/ skip) & &\checkmark & &\checkmark &    39.79 &74.65 & 56.53 & 19.94 & 36.44 & 45.99 \\

\hline
squashVGG512 C3D512 & &\checkmark&\checkmark&\checkmark &40.12 & 73.95 & 56.73 & 24.25& 36.81 & 47.27 \\
squashVGG512 C3D512 (train w/ skip) & &\checkmark&\checkmark&\checkmark & \bf{40.45} & \bf{74.16} & \bf{58.78} & \bf{23.85} & \bf{37.72} & \bf{46.72} \\

\hline
\end{tabular}
\end{center}
\caption{Detection results of different variants of trackNet on UA-DETRAC dataset (evaluated using COCO API). The average precision (AP)(\%) rates are reported under different settings (\ie IoU thresholds; bounding box area). All TrackNet variants reported here used 300 top proposals during the test except the one indicated with 2000 proposals.}
\label{coco_AP}
\end{table*}

 \begin{figure}[t]
\begin{center}
  \includegraphics[width=1.0\linewidth]{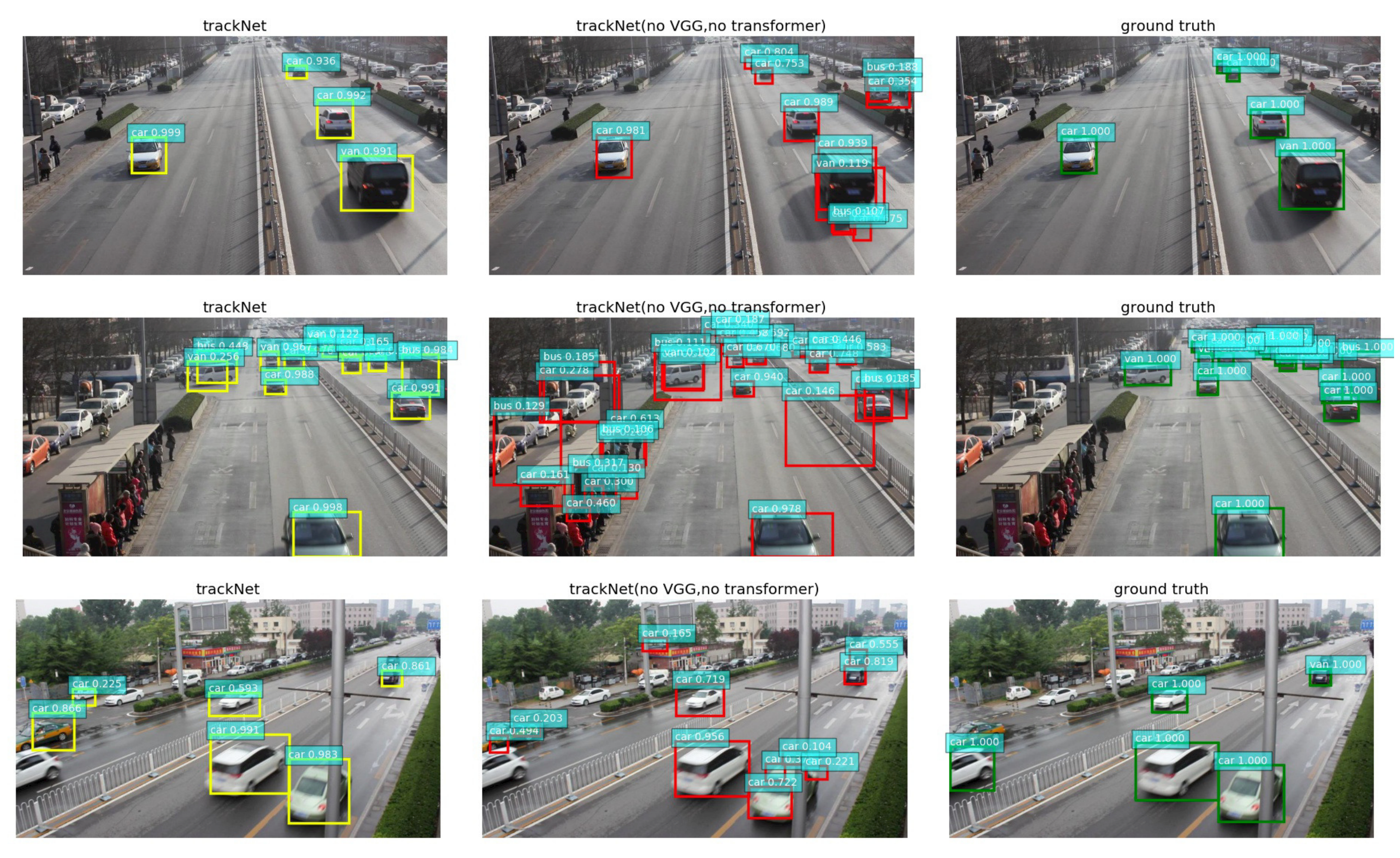}
   \caption{Result comparison for trackNet$^*$ and trackNet without VGG or transformer. Note that some false positive detections are eliminated after adding VGG branch and the spatial transformer. (Left column: trackNet$^*$, middle column: trackNet without VGG or transformer, right column: ground truth.)}
\end{center}
\label{box_video}
\end{figure}

\begin{table*}[t]
\begin{center}
\centering
\begin{tabular}{|c| c|c| c| c|c| c| c|c| c| c|c|}
\hline

&&& & &\multicolumn{3}{ c |}{ AR@IoU:} &\multicolumn{3}{ c |}{AR num maxDets:} \\
&$T_\theta(G)$ & VGG &C3D & LP  &0.1 & 0.3 & 0.5&  1 &  10 & 100 \\
\hline

 C3D head only  &&&&&&&&& & \\
 (no VGG, no transformer, predict all) &&&\checkmark& &43.59 &27.64&10.46& 8.34 & 13.34 & 13.76  \\
\hline
 C3D head only w/ LP  &&&&&&&&& & \\
 (no VGG, no transformer, interpolate)& &&\checkmark&\checkmark& 55.59&49.41&36.62& 15.27 & 25.88 & 26.43  \\
\hline
  TrackNet (no transformer)&&\checkmark&\checkmark&\checkmark&71.47& 66.92& 48.64& 25.31 & 36.46 & 36.85  \\
\hline
 TrackNet$^*$ Left view only &  \checkmark&\checkmark&\checkmark&\checkmark&
 68.50  &64.18& 45.53&  26.88&   35.08&    35.08   \\
TrackNet$^*$ Right view only &\checkmark&\checkmark&\checkmark&\checkmark& 69.92& 63.10& 45.86& 23.14 & 35.30 & 35.66  \\
TrackNet$^*$ Frontal view only &\checkmark&\checkmark&\checkmark&\checkmark& 69.98& 67.72& 55.50 & 26.37 & 38.58 & 39.08   \\
\hline

 \rowcolor{Gray}
TrackNet$^*$&\checkmark&\checkmark&\checkmark&\checkmark &71.50 & 67.01 & 49.61& 26.11 & 36.95 & 37.26   \\
 \rowcolor{Gray}
 TrackNet$^*$(2000 ROI during test) &\checkmark&\checkmark&\checkmark&\checkmark&\bf{80.97} & 74.23 & 51.24 & 27.89 & 39.71 & 40.14  \\
\rowcolor{Gray}
TrackNet$^*$ (train w/ flipped) &\checkmark &\checkmark &\checkmark&\checkmark & 78.83 & 75.81& 57.10 & 28.85 & 41.81 & 42.34\\

 \rowcolor{Gray}
TrackNet$^*$ (train w/ skip)  &\checkmark& \checkmark&\checkmark&\checkmark & 79.44 & \bf{76.45}& \bf{58.29} & \bf{29.41} & \bf{42.25} & \bf{42.80} \\

\hline
\textit{\bf{Below: Increase squash dimension from 128 to 512}}:\\
\hline
squashVGG512noC3D& &\checkmark&& \checkmark &77.35 & 74.02 & 56.07 &28.54 & 40.96 & 41.31\\

squashVGG512noC3D (train w/ skip) & &\checkmark& &\checkmark &   \bf{81.48} & \bf{78.82} & 64.11& 30.42 & 44.76 & 45.58  \\

\hline

squashVGG512 C3D512 & &\checkmark&\checkmark&\checkmark & 80.81& 78.35 & 63.77 & 30.05 & 44.83 & 45.59\\
squashVGG512 C3D512 (train w/ skip)  & &\checkmark&\checkmark&\checkmark & 80.59 & 78.47 & \bf{65.53} & \bf{30.47} & \bf{45.25} & \bf{46.06}\\

\hline
\end{tabular}
\end{center}
\caption{The average recall (AR)(\%) rates are reported under different settings (\ie IoU thresholds; thresholds on max detections per image).}
\label{coco_AR}
\end{table*}

\subsection{Training} During training, 8 sequential video frames are randomly selected from the training set (we will talk about the ``skip frame" training trick in section \ref{skip}). We first fine-tune the VGG branch alone under the framework of faster R-CNN \cite {faster_rcnn} using the whole training set as a warmup. The proposed TrackNet is then trained with the VGG branch frozen. We also fine-tune the last convolution layer (conv5a, conv5b) in the C3D backbone. The initial learning rate was $0.001$ and was reduced by 10 times after 10K iterations. We used Adam optimizer and trained the model for total 50K iterations. We evaluate the trackNet using the standard COCO API \cite{coco}. Some visual detection and tracking results are in Figure \ref{tubes}.

\subsubsection{Training with \textbf{``Skip frames"}}
\label{skip}
In the current implementation, during training, 8 consecutive video frames are randomly selected from the training set. However, due to the fact that different vehicles possess different speeds in different views (\eg larger motion when vehicles are nearer the camera, smaller motion in the back; larger motion in side-view videos etc.), training just with ``original speeds" are not sufficient. We propose a training trick: ``skip frames" to train trackNet. During training, before fetching a new GoP, a random integer skip factor $s$ is chosen $(s\in [0,5])$. Instead of fetching a consecutive 8-frame long GoP, a GoP of sampled subsequent frames are fetched. For example, when $s=1$, frame $0,2,4,6,8,10,12,14$ are selected instead of frame $0,1,2,3,4,5,6,7$. When $s=4$, frame $13,17,21,25,29,33,37,41$ are selected instead of frame $13,14,15,16,17,18,19,20$. By randomly selecting the skip factor during training, the network is more robust to different speeds. Some evaluation results are shown in table \ref{coco_AP} and table \ref{coco_AR}.

\textbf{Object Detection Performance and Ablation Analysis}. We consider all bounding tubes generated by trackNet and evaluate all bounding boxes in each frame. Table \ref{coco_AP} and \ref{coco_AR} show the average precision (AP) and average recall (AR) rates for different evaluation conditions. The criteria for labeling a detection as a true match is stricter when one increases the IoU threshold. 

In order to understand the roles the major design components are playing, different variants of trackNet are trained and tested. In table \ref{coco_AP} and \ref{coco_AR}, we show the comparisons between trackNet without transformer, trackNet without VGG feature concatenation, trackNet without transformer or VGG in either ``predicting all" mode or ``linear interpolation (LP)" mode during TPN and the full-version trackNet$^*$. We further split the training and testing dataset based on the viewing angles into left view, right view and frontal view, and show the performances when training and testing only on the sub-datasets separately. From the table it is clear that the performances got boosted after VGG concatenation and inserting spatial transformer. The linear interpolation (LP) has conveniently served as an implicit smoothness regularization during TPN stage and improved the performance with even fewer parameters. Regarding dataset, the levels of difficulty are different for different viewing angles, for example, frontal view sub-dataset is the easiest. The proposed model is improved after we augmented the training data by horizontally flipping. We expect the proposed model will perform even better given more training data and/or using other data augmentation tricks. Training with ``skip frames" also helps to boost the performance.

Based on the observations we can see that the proposed model has a higher precision rate (less false positives) thanks to the joint appearance and motion information. A candidate tube is considered to be an object only if both spatial and motion features are strong. However, the proposed model has limited power in terms of precise localization. Some ``not-so-tight" boxes can be spotted in the result visualization, figure \ref{tubes}, which is expected given the features are from a GoP level, compared with frame level features with higher temporal resolutions.

For a video segment of $T$ frames long, the feature dimension is $512\times T$ from the VGG branch, and $512$ from the C3D branch. By squashing the feature dimension from both branches into $128$, the computation is significantly reduced, making the proposed model more efficient. On the other hand, this huge dimension reduction may have lost some feature details, making the model less accurate in locating objects precisely. Indeed, the performance gets higher if using larger feature dimensions, see the second part ``increase squash dimension from 128 to 512" in both \ref{coco_AP} and \ref{coco_AR}. For example, the model named \textbf{``squashVGG512 C3D512"} has squashed feature dimensions of both feature streams into 512, which is a direct comparison with the trackNet$^*$ where both feature streams are squashed into 128. The mAP gets improved from $37.47\%$ to $40.45\%$ after increasing the squashing dimension from 128 to 512. 

\begin{figure*}[!ht]
\begin{center}
  \includegraphics[width=1.0\textwidth]{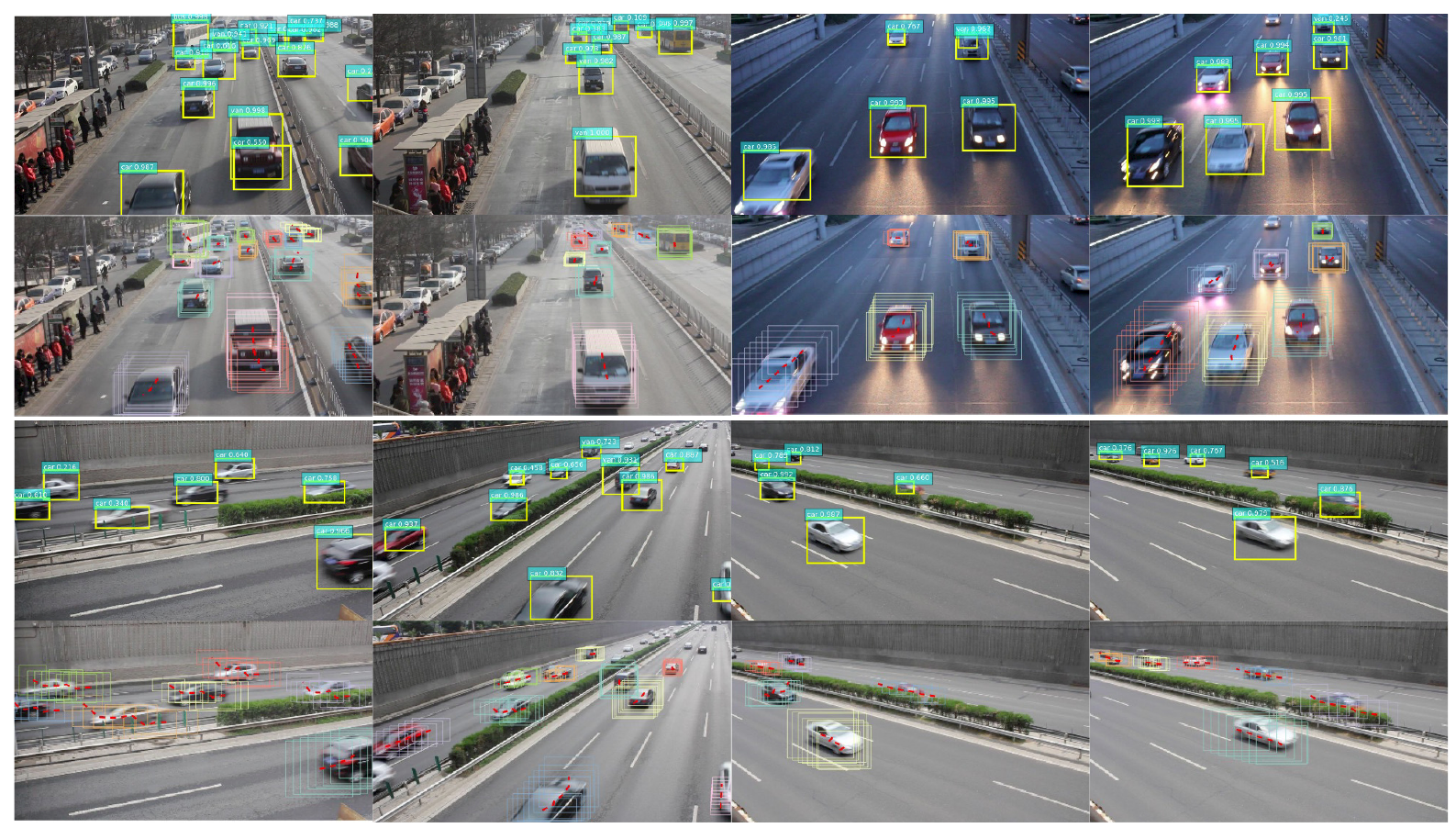}
   \caption{More examples of the predicted bounding tubes. The 1st and the 3rd row show the middle bounding boxes on the middle frame, while the 2nd and the 4th row show the bounding tubes with the centroid tracklets.}
    \label{tubes2}
\end{center}
\end{figure*}

\section{Conclusions}
We present the trackNet, which can detect and track multiple objects in videos jointly by generating bounding tubes. Utilizing the spatial-temporal features extracted by a 3D convolutional neural network in addition to the spatial features from the VGG network, the trackNet generates tube proposals and further classifies them and refines their locations. TrackNet consists of three stages: $(1)$ feature extraction and spatial transformation, $(2)$ Tube proposal network(TPN), and $(3)$post-TPN classification and refinement. We explored several ways to perform tube proposal and offset regression. TrackNet was trained and tested on the challenging traffic video dataset UA-DETRAC and achieved very promising results. In future work, we would like to improve trackNet in terms of more precise localization. Pooling features from multiple scales in spatial and temporal domain will be tested and linear interpolation structure will be relaxed to allow more complex motion patterns. \footnote{We provide two videos \url{https://drive.google.com/drive/folders/1mpbfOq1ESJ4OtXSlerGflkbxpR_6CuOX?usp=sharing}:1. Detected bounding tubes and tracklets are shown in the video \textit{trackNet\_tube.avi}. Different test videos with different viewing angles, weather conditions, lighting conditions are shown.  2. We also show the per-frame bounding boxes in the video \textit{trackNet\_box.avi} as illustrated in figure \ref{box_video}. TrackNet*, trackNet without VGG or transformer are compared to show the improvement.}

\ifCLASSOPTIONcaptionsoff
  \newpage
\fi

\bibliographystyle{IEEEtran}
\bibliography{egbib}

\end{document}